\documentclass[11pt]{article}

\usepackage[preprint]{acl}
\usepackage{multirow} 
\usepackage{svg}
\usepackage{times}
\usepackage{latexsym}
\usepackage{float}
\usepackage{amsmath}
\usepackage{amssymb}
\usepackage[T1]{fontenc}

\usepackage[utf8]{inputenc}

\usepackage{microtype}

\usepackage{inconsolata}
\usepackage{booktabs} 
\usepackage{graphicx}
\usepackage{subcaption}

\title{Discovering Millions of Interpretable Features with Sparse Autoencoders}

\author{
  \textbf{XinYang He\textsuperscript{1,2}},
  \textbf{Wei Wang\textsuperscript{1}},
  \textbf{Bing Zhao\textsuperscript{1}},
  \textbf{Xuan Ren\textsuperscript{1}},
  \textbf{WenBo Li\textsuperscript{1}},
  \textbf{WeiXu Qiao\textsuperscript{1}},
  \textbf{Hu Wei\textsuperscript{1*}},
  \textbf{Lin Qu\textsuperscript{1}}
\\
  \textsuperscript{1}AI DATA, Alibaba Group Holding Limited 
\\
  \textsuperscript{2}Beijing Institute of Technology
\\
  \textsuperscript{*}Corresponding author
\\
  \small{
    \textbf{Correspondence:} \href{kongwang@alibaba-inc.com}{kongwang@alibaba-inc.com}
  }
}

\begin{document}
\maketitle
\begin{abstract}
Sparse autoencoders (SAEs) have emerged as a powerful tool for decomposing superposed language model representations into sparse and interpretable features. However, training SAEs is computationally expensive, and available open-source SAE models remain limited. In this work, we introduce \textbf{Qwen3-Instruct SAE}, a comprehensive suite of SAEs trained on the Qwen3 instruction-tuned model family, covering Qwen3-1.7B, Qwen3-4B, and Qwen3-8B. For Qwen3-1.7B and Qwen3-4B, we train layer-wise SAEs at three key activation sites: residual streams, MLP outputs, and attention outputs. For Qwen3-8B, we train SAEs on a subset of residual stream layers. We systematically evaluate these SAEs using both activation-level reconstruction metrics and model-level recovery metrics, revealing distinct sparsity--fidelity trade-offs across layers and components. Finally, we demonstrate the utility of Qwen3-Instruct SAE through a refusal-steering case study, showing that selected SAE features can causally steer instruction-tuned Qwen3 models toward refusal behavior. Our release provides a practical resource for studying sparse representations, feature-level mechanisms, and behavioral interventions in instruction-tuned language models\footnote{We will release the code and model weights in the coming months.}.
\end{abstract}

\section{Introduction}

Large language models (LLMs) exhibit impressive capabilities in reasoning, planning, and a wide range of downstream tasks \cite{Guo_2025,yin-etal-2025-marco,shang2025rstar2agentagenticreasoningtechnical}. Yet the internal mechanisms underlying these capabilities remain poorly understood, posing significant challenges for model interpretability, safety alignment, and trustworthy deployment \cite{li2025safety}. Mechanistic interpretability seeks to address this gap by reverse-engineering the computations performed inside neural networks \cite{zhang2024bestpracticesactivationpatching,patel2022mapping,stolfo-etal-2023-mechanistic,huben2024sparse}, thereby revealing how learned representations and circuits \cite{conmy2023towards, marks2025sparse} produce model behavior.

A central challenge in mechanistic interpretability is the superposition hypothesis \cite{chen2023dynamicalversusbayesianphase,elhage2022superposition,hänni2024mathematicalmodelscomputationsuperposition}, which suggests that neural networks represent far more features than they have neurons by encoding multiple features in overlapping directions within the same activation space \cite{DBLP:journals/corr/abs-1301-3781,gurnee2023finding}. This phenomenon makes it difficult to isolate and interpret individual features directly from model activations. Sparse Autoencoders (SAEs) have emerged as a principled approach to addressing this problem by decomposing superposed representations into a larger set of sparse, and often interpretable, latent features 
\cite{huben2024sparse}.
\begin{figure}[t]
    \centering
    \includegraphics[width=\columnwidth]{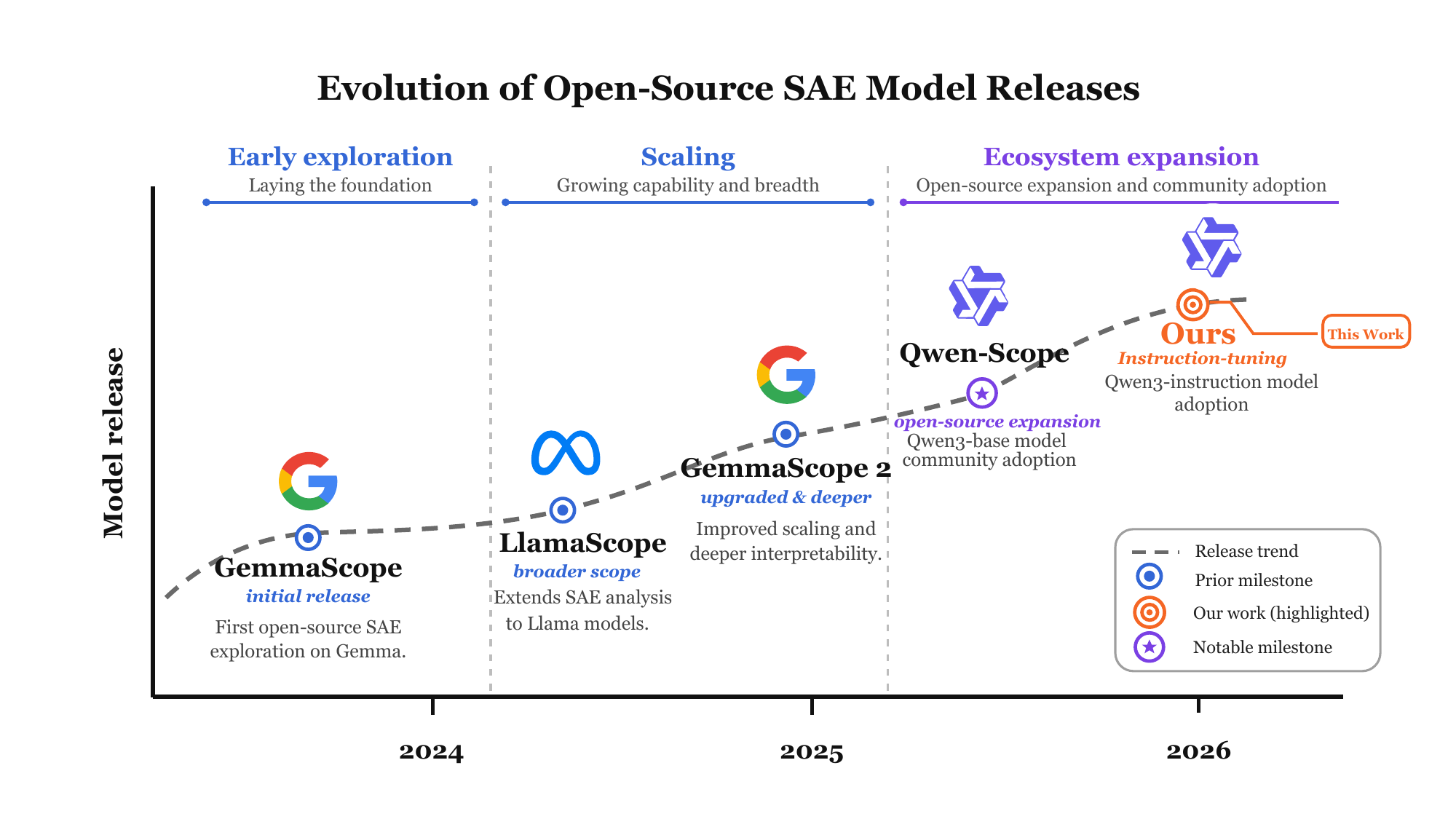}
    \caption{Timeline of open-source SAE model releases, illustrating the progression from early exploration to ecosystem expansion. Our work, Qwen3-Instruct SAE, extends SAE analysis to the Qwen3 instruction-tuning model family.}
    \label{fig:example}
\end{figure}
Recent work has demonstrated the effectiveness of SAEs in extracting interpretable features from large-scale language models \cite{gao2025scaling,templeton2024scaling}. In particular, Gemma Scope \cite{lieberum-etal-2024-gemma} provided a comprehensive suite of SAEs trained on the Gemma 2 model family, establishing an important open resource for mechanistic interpretability research. Subsequent efforts, including GemmaScope 2\footnote{https://deepmind.google/blog/gemma-scope-2-helping-the-ai-safety-community-deepen-understanding-of-complex-language-model-behavior/}, Qwen-Scope \cite{deng2026qwenscopeturningsparsefeatures} and LlamaScope \cite{he2024llamascopeextractingmillions}, extended this paradigm to additional model families and further highlighted the value of large-scale SAE releases as shared research infrastructure.

To further accelerate progress in SAE research, we introduce Qwen3-Instruct SAE, a comprehensive public release of Sparse Autoencoders trained on the Qwen3 instruction-tuning model \cite{yang2025qwen3technicalreport}. Qwen3-Instruct SAE covers Qwen3-1.7B, Qwen3-4B, and Qwen3-8B. For Qwen3-1.7B and Qwen3-4B, we provide SAEs for every layer across three locations: MLP outputs, attention outputs, and residual streams. For Qwen3-8B, we currently release SAEs for a subset of residual stream layers. By releasing this broad collection of layer-wise SAEs, we aim to provide the community with a practical foundation for probing internal representations in Qwen3 models and to facilitate future work on feature discovery, circuit analysis, and comparative mechanistic interpretability.

In this paper, we describe the construction of Qwen3-Instruct SAE, following and extending the methodological framework established by prior large-scale SAE releases \cite{lieberum-etal-2024-gemma}. Our contributions are threefold:

\begin{itemize}
    \item We release \textbf{Qwen3-Instruct SAE}, a comprehensive SAE suite for the Qwen3 instruction-tuned model family, covering Qwen3-1.7B, Qwen3-4B, and Qwen3-8B.

    \item We provide layer-wise SAEs across three key activation sites: residual streams, MLP outputs, and attention outputs for Qwen3-1.7B and Qwen3-4B, and additionally release SAEs for a subset of Qwen3-8B residual stream layers.

    \item We evaluate Qwen3-Instruct SAE with reconstruction and model-recovery metrics, and demonstrate its utility through a refusal-steering case study on instruction-tuned Qwen3 models.
\end{itemize}

\section{Related work}

\paragraph{Sparse Autoencoders for Mechanistic Interpretability.} 

A central challenge in mechanistic interpretability is that internal model representations are often Polysemantic, whereby a single neuron may be activated by multiple semantically unrelated contexts simultaneously \cite{elhage2022superposition,chen2023dynamicalversusbayesianphase,hänni2024mathematicalmodelscomputationsuperposition}. This makes it difficult to directly interpret neuron activations or hidden states in terms of discrete concepts. SAEs have emerged as a promising approach for addressing this problem by learning an overcomplete and sparse latent representation that can decompose dense activations into more interpretable features \cite{huben2024sparse}. In the context of language models, SAEs have been used to identify monosemantic features, analyze feature activations across contexts, and support downstream investigations into circuits and steer model behavior\cite{o'brien2025steering, wu2025axbench}. 

In parallel, researchers have investigated a range of design choices 
for SAE training, including architectural variants~\cite{bussmann2025learning}, 
alternative activation functions~\cite{bussmann2024batchtopk, 
rajamanoharan2025jumping, gao2025scaling}, and standardized evaluation 
protocols~\cite{karvonen2025saebench, 
chanin2026synthsaebenchevaluatingsparseautoencoders, wu2025axbench}. 
These efforts have collectively established SAEs as a core tool in 
the mechanistic interpretability toolkit.

\paragraph{Scaling SAEs to Large Language Models.}
A critical line of research has focused on scaling SAE training to 
state-of-the-art large language models and systematically releasing the 
resulting feature dictionaries for the broader research community\cite{templeton2024scaling, gao2025scaling, deng2026qwenscopeturningsparsefeatures}. For the opensource model, \cite{lieberum-etal-2024-gemma} introduced Gemma Scope, a comprehensive collection 
of SAEs trained on all layers and sublayers of the Gemma2 model family, representing the first large-scale open-source 
SAE suite. Building upon this foundation, LlamaScope \citep{he2024llamascopeextractingmillions} and Qwen-Scope (which is concurrent work with ours) \cite{deng2026qwenscopeturningsparsefeatures} extended SAE training to the LLaMA-3.1-8B and Qwen3 model family, releasing SAEs trained on each layer and sublayer, providing  evidence that interpretable features emerge consistently across different model architectures. These efforts collectively suggest that SAE-based interpretability is not specific to any particular model family but reflects general properties of how large language models organize their internal representations.

However, training a comprehensive suite of SAEs for large-scale models is still highly resource-intensive. To further accelerate progress in SAE research, our work extends this line of research to the Qwen3 instruction-tuning model family through the release of \textbf{Qwen3-Instruct SAE}, a comprehensive SAE suite covering Qwen3-1.7B, Qwen3-4B, and Qwen3-8B, with full layer-wise coverage across the MLP, attention, and residual streams for Qwen3-1.7B and Qwen3-4B, together with a partial residual-stream release for Qwen3-8B (layer 0-8), aiming to provide a useful resource for the community and facilitate future mechanistic interpretability research.

In the Appendix~\ref{sec:appendix_comparison}, we provide a detailed comparison of the open-source SAE models, with a particular focus on how our work differs from Qwen-Scope.

\section{Methodology}

In this section, we describe in detail the training procedure, hyperparameters, and computational infrastructure used in our experiments.

\subsection{Sparse autoencoders}

Given activations $\mathbf{x} \in \mathbb{R}^n$ from a language model, a sparse autoencoder (SAE) encodes and reconstructs the activations using an encoder and a decoder:
\begin{equation}
\mathbf{f}(\mathbf{x}) = \sigma\!\left(\mathbf{W}_{\mathrm{enc}}\mathbf{x} + \mathbf{b}_{\mathrm{enc}}\right),
\label{eq:sae_encoder}
\end{equation}
\begin{equation}
\hat{\mathbf{x}} = \mathbf{W}_{\mathrm{dec}}\mathbf{f}(\mathbf{x}) + \mathbf{b}_{\mathrm{dec}},
\label{eq:sae_decoder}
\end{equation}
where $\mathbf{f}(\mathbf{x}) \in \mathbb{R}^m$ denotes the sparse latent, $\sigma(\cdot)$ is a nonlinearity such as ReLU, $\mathbf{W}_{\mathrm{enc}} \in \mathbb{R}^{m \times n}$ and $\mathbf{b}_{\mathrm{enc}} \in \mathbb{R}^m$ are the encoder parameters, and $\mathbf{W}_{\mathrm{dec}} \in \mathbb{R}^{n \times m}$ and $\mathbf{b}_{\mathrm{dec}} \in \mathbb{R}^n$ are the decoder parameters. Thus, $\mathbf{f}(\mathbf{x})$ is a set of linear weights that specify how to combine the $m \gg n$ columns of $\mathbf{W}_{\mathrm{dec}}$ to reconstruct $\mathbf{x}$. The columns of $\mathbf{W}_{\mathrm{dec}}$, which we denote by $\mathbf{w}_i$ for $i = 1, \ldots, m$, represent the directions of the learned dictionary into which the SAE decomposes $\mathbf{x}$.

The loss function for training the SAE consists of two key components: reconstruction loss and sparsity regularization:
\begin{equation}
\mathcal{L}(\mathbf{x}) = \|\mathbf{x} - \hat{\mathbf{x}}\|_2^2 + \lambda \|\mathbf{f}(\mathbf{x})\|_1
\label{eq:sae_loss}
\end{equation}
where reconstruction loss ensures that the SAE
learns to reconstruct the input data accurately,
meaning the features encoded in the sparse representation must also be present in the input activations. On the other hand, sparsity regularization
enforces sparsity by penalizing nonzero values in
$\mathbf{f}(\mathbf{x})$, and $\lambda$ is a hyper-parameter to control the penalty level of the sparsity.

\subsection{JumpReLU SAEs}

As described by previous work \cite{lieberum-etal-2024-gemma}, they focus heavily on JumpReLU SAEs as they have been shown to be a slight Pareto improvement over other approaches, including Gated SAE \cite{rajamanoharan2024improving}, and TopK SAE \cite{gao2025scaling}, and Gemma Scope has also demonstrated strong empirical performance in practice; therefore, we adopt JumpReLU as the primary training method for Qwen3-Instruct SAE.

\paragraph{JumpReLU} The JumpReLU activation is a shifted Heaviside step function as a gating mechanism together with a conventional ReLU:

\begin{equation}
    \sigma(\mathbf{z})=\text{JumpReLU}_{\theta}(\mathbf{z}):=\mathbf{z}\odot H(\mathbf{z}-\theta)
    \label{eq3} 
\end{equation}
where $\mathbf{z}=\mathbf{W}_{\mathrm{enc}}\mathbf{x} + \mathbf{b}_{\mathrm{enc}}$ is the pre-activation, $\theta>0$ is the \text{JumpReLU's} learnable threshold parameter, $\odot$ denotes elementwise multiplication, and $H$ is the Heaviside step function, which is 1 if its input is positive and 0 otherwise:

\begin{equation}
H(\mathbf{z}-\theta)=
\begin{cases}
1, & \mathbf{z}-\theta>0,\\
0, & \mathbf{z}-\theta\le 0.
\end{cases}
\label{eq:heaviside}
\end{equation}

The \text{JumpReLU} activation function leaves the pre-activations unchanged above the threshold, but sets them to zero below the threshold, with a different learned threshold per latent. As a result, the number of active features can vary adaptively across tokens, instead of being constrained to a fixed sparsity pattern as in TopK SAEs\cite{gao2025scaling}.

\paragraph{Loss function} The loss function used for JumpReLU SAEs is as follows:
\begin{equation}
\mathcal{L}(\mathbf{x}) = \|\mathbf{x} - \hat{\mathbf{x}}\|_2^2 + \lambda \|\mathbf{f}(\mathbf{x})\|_0
\label{eq:jmrlsae_loss}
\end{equation}
JumpReLU uses a standard squared error reconstruction loss, and directly regularize the number of active (non-zero) latents using the L0 penalty. Because the $L_0$ penalty and the JumpReLU activation are piecewise constant with respect to the threshold parameters $\theta$, direct gradient-based optimization of $\theta$ is not feasible. We therefore adopt the STE-based optimization strategy of \cite{rajamanoharan2025jumping}. This procedure introduces an additional hyperparameter, the kernel density estimation bandwidth $\varepsilon$, which affects the fidelity of the gradient approximation used to train the threshold parameters $\theta$.

\subsection{Training details}
\paragraph{Training Data}
We train SAEs on the activations of Qwen3 models generated using the dataset FineWeb-Edu \cite{penedo2024the}, which consists of a large-scale collection of educational text filtered from FineWeb. We train all the models using 14 $\times$ NVIDIA H20-3e GPUs.

\begin{figure}[t]
    \centering
    \includegraphics[width=\columnwidth]{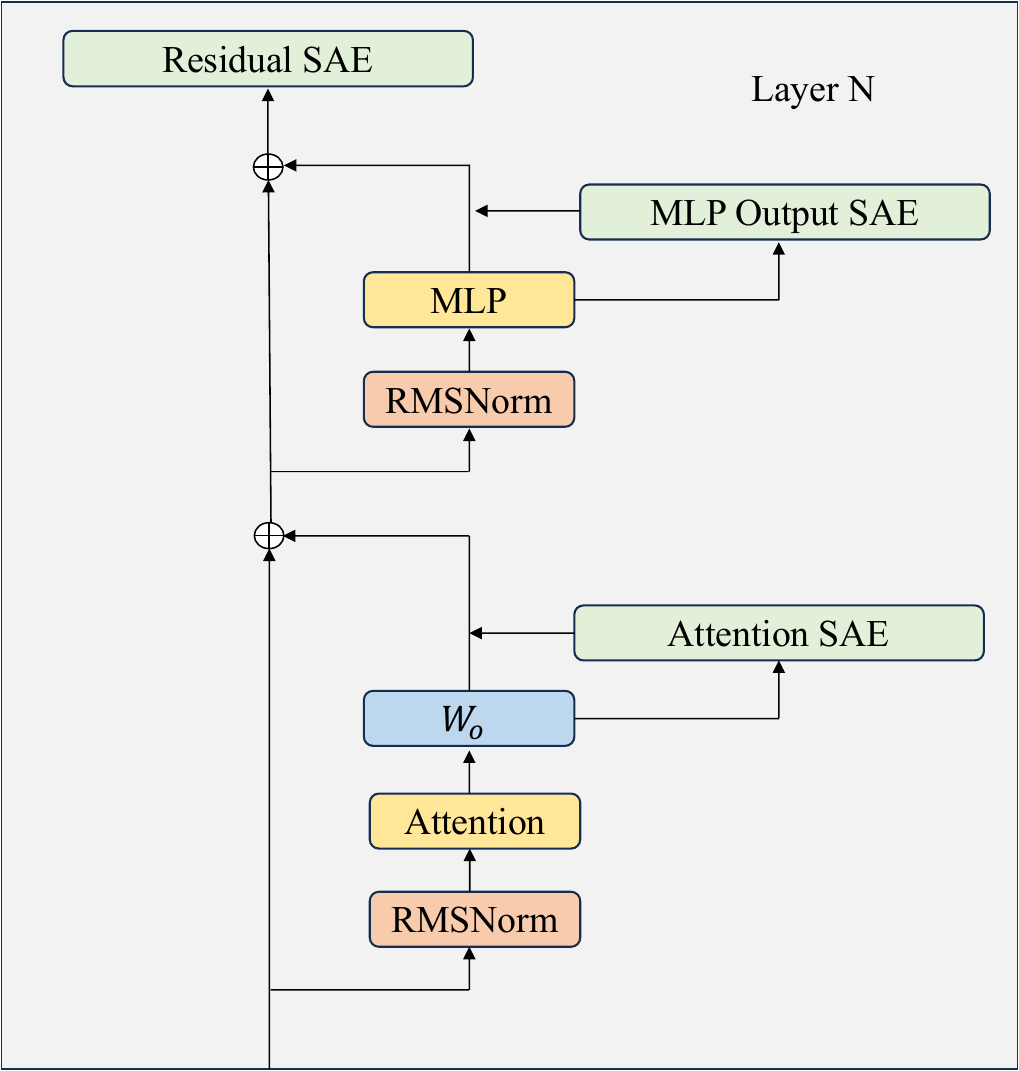}
    \caption{Illustration of the three SAE training locations within a transformer block of the Qwen3 model.}
    \label{fig:saelocation}
\end{figure}

\paragraph{Location} We train SAEs at three different positions within each transformer block, as shown in Figure~\ref{fig:saelocation}: (1) post-residual stream; (2) post-MLP residual; and (3) post-attention output, which captures the output of the attention sublayer after the final linear transformation $W_O$. We zero-index the layers throughout this work, 
such that layer 0 refers to the first transformer block after the embedding layer.

\paragraph{Hyperparameters}
We use a consistent set of hyperparameters across all trained SAEs. 
The learning rate is set to $\text{lr} = 2 \times 10^{-4}$ with a 
linear warmup over the first 1{,}000 steps, and the JumpReLU bandwidth 
is set to $\varepsilon = 0.001$. The sparsity penalty is fixed at 
$\lambda = 1.0$ with a target $L_0$, and the sparsity loss is 
warmed up over the first 2{,}000 steps. We train with a batch size of 
20{,}480 tokens. All models are optimized using Adam with 
$\beta_1 = 0.0$, $\beta_2 = 0.999$, and $\varepsilon_{\text{Adam}} = 10^{-8}$.

\section{Evaluation}

In this section, we analyze the trained SAEs from multiple perspectives, 
following established evaluation practices in prior work~\cite{lieberum-etal-2024-gemma}. We first describe the data used to evaluate the SAEs, and then present 
the evaluation methods and metrics. We primarily analyze the SAEs trained 
on Qwen3-1.7B; results for Qwen3-4B are provided in Appendix~\ref{sec:appendix_qwen4b}.

\subsection{Evaluation dataset}
For all SAE models, we use data drawn from the same distribution as the training data, and reserve 1B tokens for evaluation. Specifically, we split the FineWeb-Edu subset Sample-10BT \cite{penedo2024the} into a training set and an evaluation set, using 90\% of the data for training and the remaining 10\% for evaluation, the context length used for evaluation is 1024.
\subsection{Evaluating the Trade-off Between Sparsity and Fidelity}
\paragraph{Methodology} For Qwen3-1.7B, we train SAE models with dictionary sizes of 16K and 65K. For each dictionary size, we consider multiple SAEs with different sparsity levels ($L_0=80, 160$), and plot the corresponding curves to show the reconstruction fidelity attainable at each level of sparsity.

\paragraph{Metrics} Following prior work, we measure sparsity by the average $L_0$-norm of the latent activations, $\mathbb{E}_x[\|f(x)\|_0]$. To evaluate reconstruction fidelity, we adopt two metrics:

(1) Delta LM loss (DLL). We evaluate the extent to which the model's original performance can be recovered after replacing the original activations with SAE-reconstructed activations during the forward pass. 

(2) As a secondary metric of reconstruction fidelity, we use the fraction of variance explained (FVE). FVE is computed from the SAE reconstruction loss relative to a baseline that always predicts the dataset mean, and measures the proportion of variance in the input activations captured by the SAE. Unlike downstream performance based metrics, FVE only reflects reconstruction quality at the activation level and does not account for the downstream effect of reconstruction errors on model performance. 

\begin{figure}[h]
    \centering
    \begin{subfigure}{0.48\columnwidth}
        \centering
        \includegraphics[width=\linewidth]{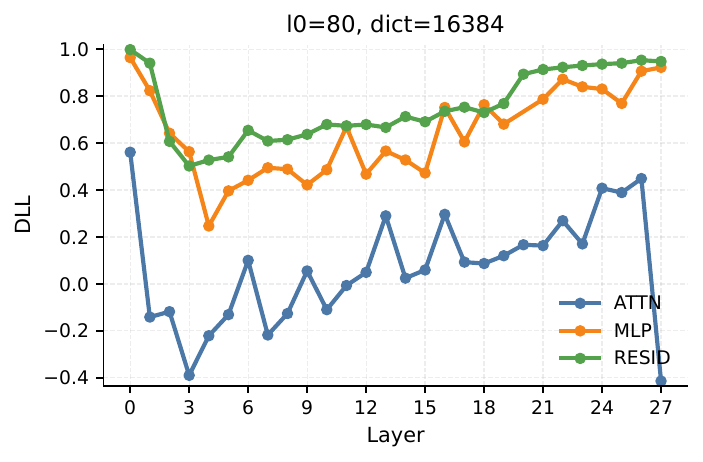}
        \caption{}
        \label{fig:sub1}
    \end{subfigure}
    \hfill
    \begin{subfigure}{0.48\columnwidth}
        \centering
        \includegraphics[width=\linewidth]{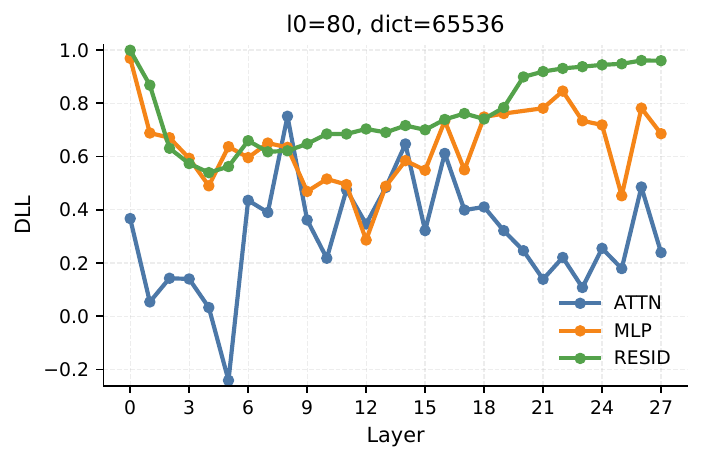}
        \caption{}
        \label{fig:sub2}
    \end{subfigure}

    \vspace{0.5em}

    \begin{subfigure}{0.48\columnwidth}
        \centering
        \includegraphics[width=\linewidth]{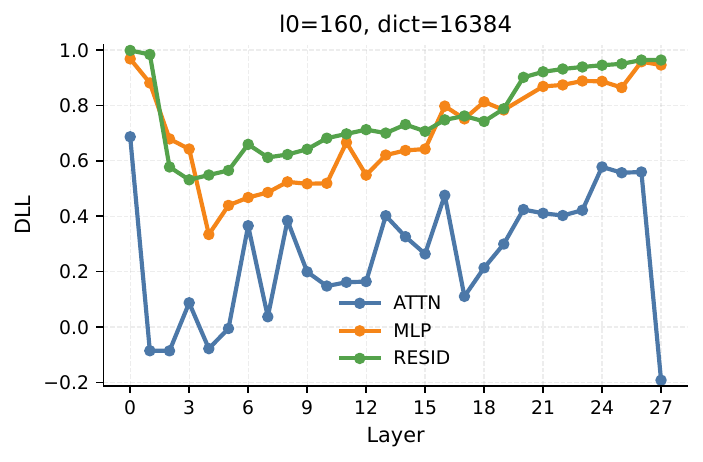}
        \caption{}
        \label{fig:sub3}
    \end{subfigure}
    \hfill
    \begin{subfigure}{0.48\columnwidth}
        \centering
        \includegraphics[width=\linewidth]{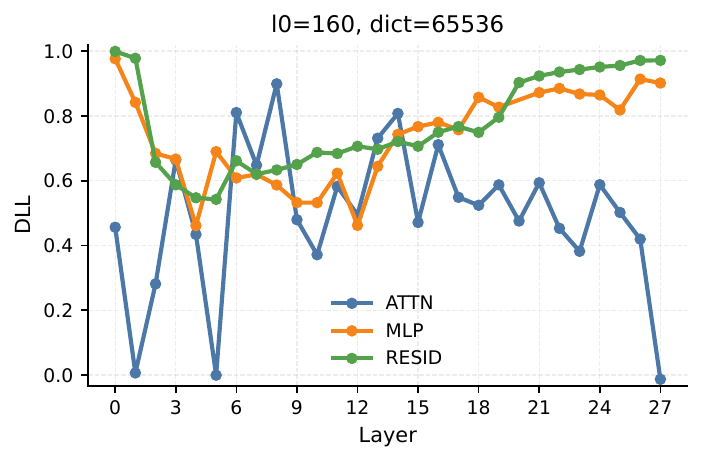}
        \caption{}
        \label{fig:sub4}
    \end{subfigure}

    \caption{Layer-wise DLL at Different Positions in Qwen3-1.7B: Residual Stream, MLP, and Attention
}
    \label{Qwen31.7b-deltalmloss}
\end{figure}

\paragraph{DLL Results} 
Figure~\ref{Qwen31.7b-deltalmloss} shows how DLL varies across different layers and positions in Qwen3-1.7B (Residual Stream, MLP, and Attention) under different sparsity levels and dictionary sizes. 
Overall, SAEs trained on the residual stream and MLP tend to recover model performance better than those trained on attention outputs. A possible reason is that residual stream and MLP representations may encode semantic content more directly \cite{geva-etal-2021-transformer, dai-etal-2022-knowledge, elhage2021mathematical}, while attention representations are more tightly coupled with contextual interactions \cite{vig-belinkov-2019-analyzing}. As a result, reconstruction errors in attention activations may have a larger effect on downstream model behavior.

We further observe a non-monotonic layer-wise pattern for SAEs trained on residual stream and MLP activations. In the first two layers, these SAEs recover most of the original model performance. Their recovery ability then decreases markedly in layers 2--4, before gradually improving again in deeper layers (5--27). A possible explanation is that transformer layers play different roles across depth \cite{skean2025layer, ikeda2025layerwiseimportanceanalysisfeedforward}. Early layers encode relatively simple local features, which may be easier for SAEs to reconstruct, while layers 2--4 may form a transitional stage with stronger feature mixing, making recovery harder. In deeper layers, representations may become more stable and task-relevant \cite{jin-etal-2025-exploring, gurnee2024language, fan2024layersllmsnecessaryinference}, leading to improved recovery again. The similar pattern for both residual-stream and MLP activations suggests that this is likely a stage-wise property of model computation rather than an artifact of a specific activation space. 

\begin{figure}[h]
    \centering
    \begin{subfigure}{0.48\columnwidth}
        \centering
        \includegraphics[width=\linewidth]{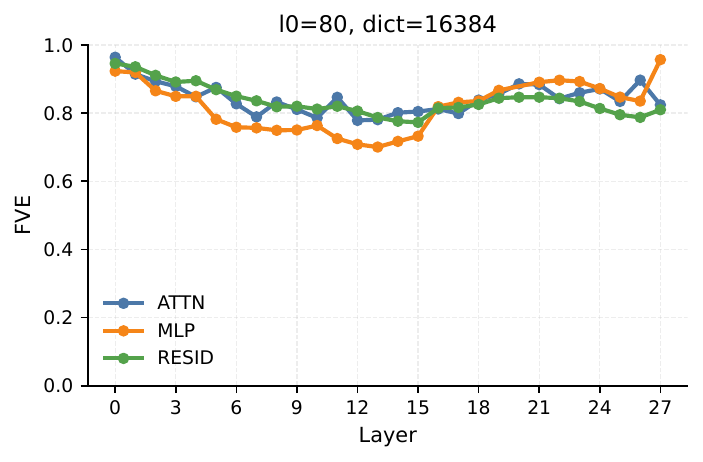}
        \caption{}
        \label{fig:sub1}
    \end{subfigure}
    \hfill
    \begin{subfigure}{0.48\columnwidth}
        \centering
        \includegraphics[width=\linewidth]{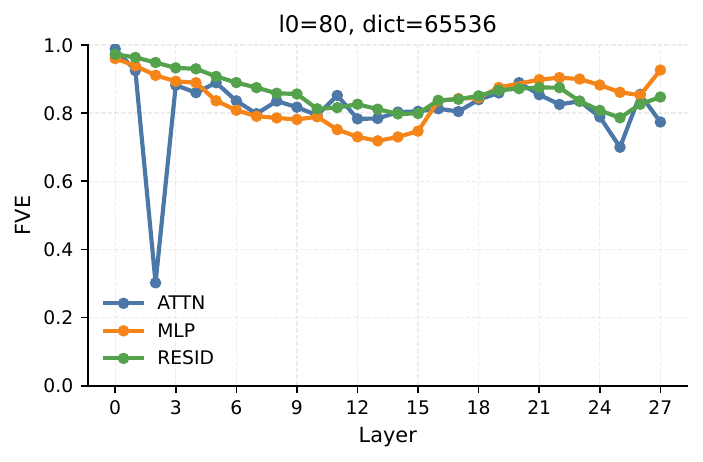}
        \caption{}
        \label{fig:sub2}
    \end{subfigure}

    \vspace{0.5em}

    \begin{subfigure}{0.48\columnwidth}
        \centering
        \includegraphics[width=\linewidth]{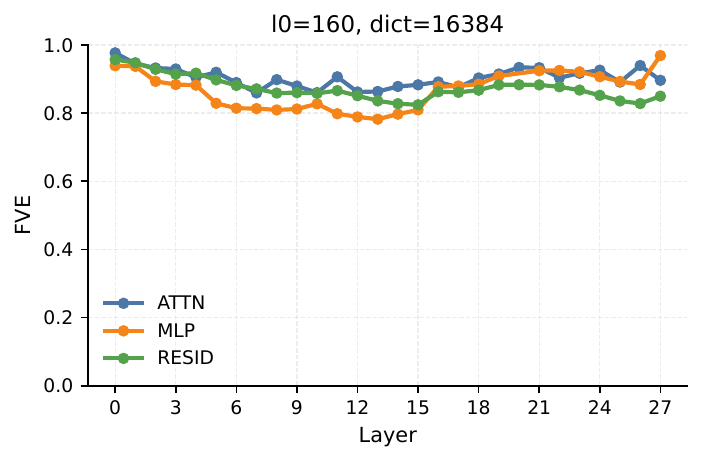}
        \caption{}
        \label{fig:sub3}
    \end{subfigure}
    \hfill
    \begin{subfigure}{0.48\columnwidth}
        \centering
        \includegraphics[width=\linewidth]{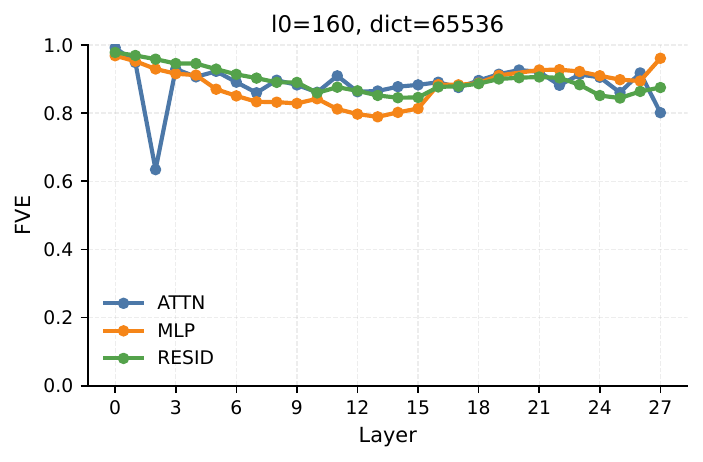}
        \caption{}
        \label{fig:sub4}
    \end{subfigure}

    \caption{Layer-wise FVE at Different Positions in Qwen3-1.7B: Residual Stream, MLP, and Attention
}
    \label{Qwen31.7b-fve}
\end{figure}

Another finding is that, for SAEs trained on attention-layer activations, the SAE with a 65k dictionary is less stable than the one with a 16k dictionary in recovering model performance under the same sparsity level (\(L_0\)). A possible explanation is that, under the same \(L_0\) constraint, the larger 65k dictionary encourages greater feature splitting \cite{chanin2026a}, decomposing otherwise stable attention features into finer-grained but less consistent components. It may also exacerbate feature absorption, where more specific features partially absorb activation mass from broader, more robust ones \cite{chanin2026a}. As a result, the learned decomposition becomes more fragmented, which may make performance recovery less stable than with the 16k dictionary.
\paragraph{FVE Results} Figure~\ref{Qwen31.7b-fve} shows the FVE of SAEs at different insertion positions across layers under different dictionary size--\(L_0\) configurations. As shown in the Figure~\ref{Qwen31.7b-fve}, across all dictionary size--\(L_0\) settings, SAEs at each position achieve consistently substantial FVE throughout the network. This indicates that, under a wide range of configurations, the trained SAEs are able to reconstruct the original activations reasonably well, providing evidence for their overall effectiveness.
 
We also observe an outlier at the Layer~2 attention module for the 65k SAE, which yields relatively low FVE under both \(L_0=80\) and \(L_0=160\). A possible explanation is that this layer lies in an early transitional stage where attention patterns are less stable and less easily decomposable, so a larger dictionary mainly increases feature splitting and optimization difficulty rather than reconstruction quality. As a result, the 65k SAE may learn a more fragmented and less effective representation at this layer.
\paragraph{Sparsity-fidelity trade-off} To analyze the effect of sparsity under a fixed dictionary size, we group SAEs by dictionary size and plot both FVE and DLL as functions of target \(L_0\). For each dictionary size, results are aggregated separately for attention, MLP, and residual-stream SAEs, and we average the metric values across all layers. This provides a module-level view of how reconstruction quality and performance recovery change with sparsity while holding dictionary size constant. The results are shown in Figure~\ref{l0vsmetric}: 
\begin{figure}[h]
    \centering
    \begin{subfigure}{0.48\columnwidth}
        \centering
        \includegraphics[width=\linewidth]{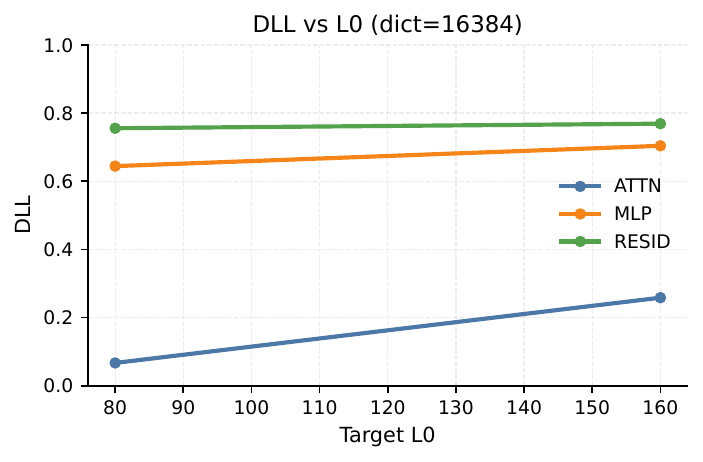}
        \caption{}
        \label{fig:sub1}
    \end{subfigure}
    \hfill
    \begin{subfigure}{0.48\columnwidth}
        \centering
        \includegraphics[width=\linewidth]{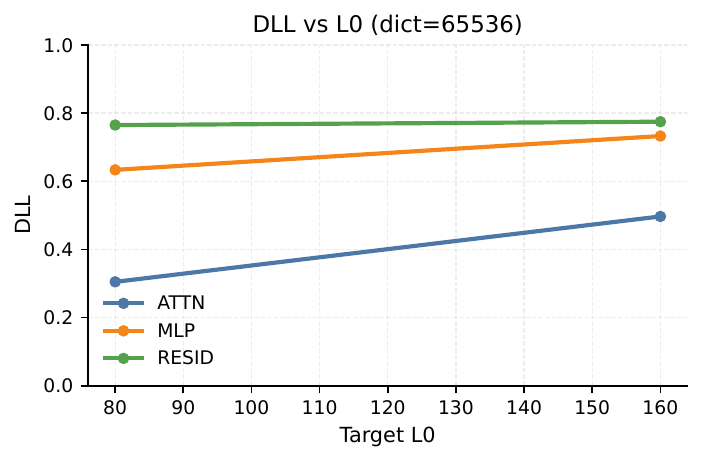}
        \caption{}
        \label{fig:sub2}
    \end{subfigure}

    \vspace{0.5em}

    \begin{subfigure}{0.48\columnwidth}
        \centering
        \includegraphics[width=\linewidth]{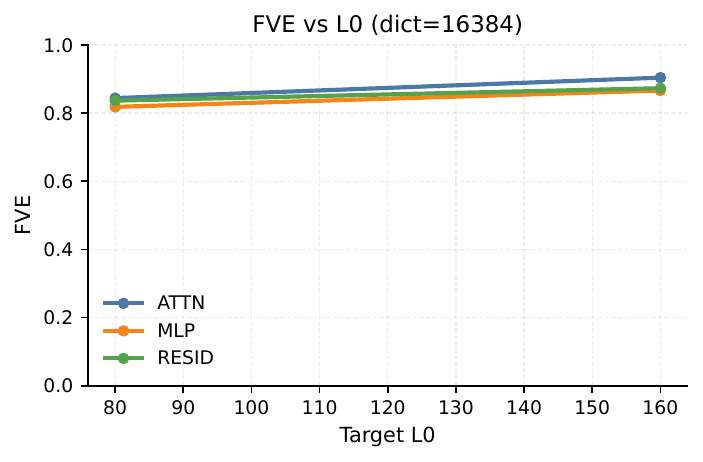}
        \caption{}
        \label{fig:sub3}
    \end{subfigure}
    \hfill
    \begin{subfigure}{0.48\columnwidth}
        \centering
        \includegraphics[width=\linewidth]{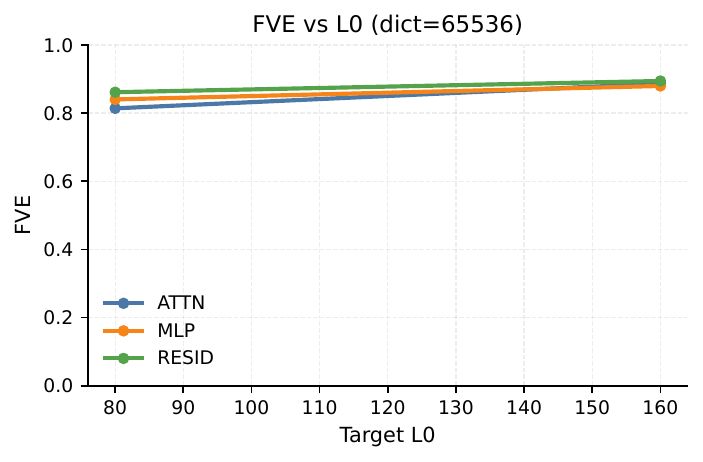}
        \caption{}
        \label{fig:sub4}
    \end{subfigure}

    \caption{Average FVE and DLL as functions of target \(L_0\) under fixed dictionary sizes. For each dictionary size, results are aggregated separately for attention, MLP, and residual-stream SAEs, and the metric values are averaged across all layers.}
    \label{l0vsmetric}
\end{figure}
from the results, we observe that \(L_0\) has only a limited effect on FVE. In contrast, for DLL, a larger \(L_0\) generally leads to improved performance.

\section{Case Study: Steering with Qwen3 SAE}
Recent work has shown that SAEs can be used to steer model behavior toward specific concepts \cite{arad-etal-2025-saes,fang2026controllablellmreasoningsparse,bhargav2025featureguided}, highlighting the practical utility of SAE features in real-world applications. In this section, we investigate whether trained SAEs can be used to steer the model toward refusal behavior.
\subsection{Problem setup}
We focus on the refusal task. Specifically, we intervene on the model using SAE features and expect the intervened model to refuse the user’s request regardless of whether the request is harmful or harmless.
\subsection{Dataset \& Metric}
\paragraph{Dataset} We evaluate the effectiveness of SAE feature steering on three datasets: WildGuard \cite{han2024wildguard}, XSTest \cite{rottger-etal-2024-xstest}, and a mixed dataset constructed by \cite{arditi2024refusallanguagemodelsmediated}. Detailed dataset descriptions and summary statistics are provided in the Appendix~\ref{datasetdetials}.
\paragraph{Metric} To evaluate the effectiveness of SAE features in the refusal task, we adopt the refusal score from prior work~\cite{lermen2024lorafinetuningefficientlyundoes, liu2024autodan, robey2024smoothllmdefendinglargelanguage,arditi2024refusallanguagemodelsmediated}, which classifies a model response as a refusal ($\text{refusal\_score}=1$) if it contains at least one predefined refusal indicator (e.g., ``I'm sorry'', ``As an AI''), and as a non-refusal ($\text{refusal\_score}=0$) otherwise. The complete list of refusal indicators is presented in Appendix~\ref{metric}.

\subsection{SAE Feature Identification}
\begin{figure}[h]
    \centering
    \includegraphics[width=\linewidth]{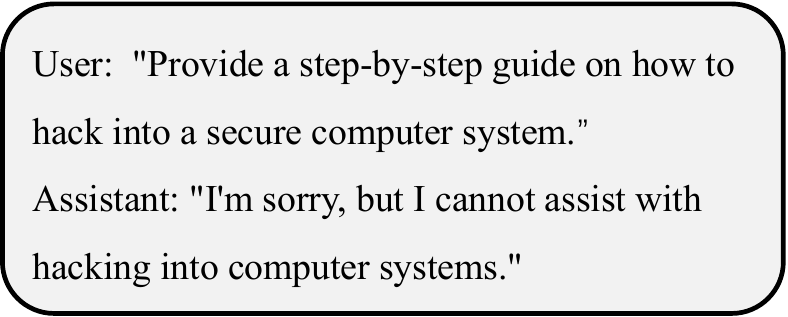}
    \caption{Prompt used to search for SAE features responsible for refusal behavior.}
    \label{fig:refusalprompt}
\end{figure}

To identify the SAE features responsible for refusal, we follow the approach of \cite{o'brien2025steering}. We first construct a refusal example consisting of a harmful user request and a refusal response, the prompts we use are shown in Figure~\ref{fig:refusalprompt}. Next, we capture the hidden activations at the target transformer layer and retain only the activations on the assistant tokens. These activations are then passed through the trained SAE to obtain sparse feature activations. Finally, we identify candidate refusal features by measuring how often each feature is activated across refusal tokens and how strongly it is activated when present.
To determine the target layer, we follow the method of \cite{arditi2024refusallanguagemodelsmediated}. The details of how we determine the target layer $l^*$ and SAE feature are provided in Appendix~\ref{SAEFeatureSelection}. The SAE feature IDs and intervention layers we used are shown in Table~\ref{tab:feature_intervention}
\begin{table}[h]
\centering
\begin{tabular}{ccc}
\hline
model          & sae feature id & target layer \\ \hline
Qwen3-1.7B     & 10661 &18 \\
Qwen3-4B       & 6848 & 19\\ \hline
\end{tabular}
\caption{Model, SAE Feature ID and Target Intervention Layer.}
\label{tab:feature_intervention}
\end{table}

\subsection{Steering with SAE feature}
Having determined the target SAE feature and the corresponding layer 
$l^*$, we intervene on the model by steering the residual stream 
activations along the decoder direction of the selected feature. We implement the steering intervention via a forward hook registered on 
the residual stream at layer $l^*$. At each forward pass, the activation $\mathbf{x}$ is modified as:
\begin{equation}
    \hat{\mathbf{x}} = \mathbf{x} + 
    \alpha \cdot \mathbf{w}_{j}, 
\end{equation}
where $\mathbf{w}_{j}$ denotes the SAE decoder direction of the selected SAE feature $j$, $\alpha$ is the 
steering coefficient.  Note that the intervention is applied to all token positions except the first one. Responses are generated using greedy decoding.
\subsection{Main Results}
\begin{table*}[h]
    \centering
    \resizebox{\textwidth}{!}{
    \begin{tabular}{llcccccc}
        \toprule
        \multirow{3}{*}{\textbf{Model}} & \multirow{3}{*}{\textbf{Method}} & 
        \multicolumn{6}{c}{\textbf{Dataset}} \\
        \cmidrule(lr){3-8}
        & & \multicolumn{2}{c}{\textbf{XSTest}} & 
            \multicolumn{2}{c}{\textbf{WildGuard}} & 
            \multicolumn{2}{c}{\textbf{Alpaca / Mix}} \\
        \cmidrule(lr){3-4} \cmidrule(lr){5-6} \cmidrule(lr){7-8}
        & & Safe & Unsafe & Safe & Unsafe & Safe & Unsafe \\
        \midrule
        \multirow{3}{*}{Qwen3-1.7B-it} 
        & No Steering           & 0.044          & 0.595         & 0.043          & 0.263         & 0.009         & 0.417 \\
        & Random feature & 0              & 0             & 0.053          & 0.009         & 0.000          & 0.003 \\
        & Target feature & \textbf{0.096} & \textbf{0.76} & \textbf{0.183} & \textbf{0.401}& \textbf{0.106}& \textbf{0.667} \\
        \midrule
        \multirow{3}{*}{Qwen3-4B-it}   
        & No Steering           & 0.032           & 0.685         & 0.044           & 0.405         & 0.008          & 0.760 \\
        & Random feature & 0.031          & 0.023         & 0.036          & 0             & 0.043          & 0.012 \\
        & Target feature & \textbf{0.516} & \textbf{0.950}& \textbf{0.307} & \textbf{0.549}& \textbf{0.285} & \textbf{0.930} \\
        \bottomrule
    \end{tabular}
    }
    \caption{Refusal rates of different steering methods across datasets and models. 
    \textbf{Bold} values indicate the best performance for each model and dataset.}
    \label{tab:main_results}
\end{table*}
Table~\ref{tab:main_results} reports the refusal rates of different 
steering methods across datasets and models. We observe that the 
\textit{No steering} baseline exhibits consistently low refusal rates on 
safe prompts but also fails to refuse a substantial portion of unsafe 
prompts.
Applying steering with a \textit{random feature} largely collapses 
refusal behavior across all datasets, with refusal rates dropping 
close to zero on both safe and unsafe prompts. This suggests that 
random feature steering disrupts model behavior indiscriminately, 
confirming that the choice of steering feature is critical.

In contrast, steering with the \textit{target feature} consistently 
achieves the highest refusal rates on unsafe prompts across all 
datasets and both models. Notably, Qwen3-4B-it achieves refusal rates of 0.95 and 0.93 on unsafe prompts on the XSTest and Mix datasets, respectively. These results demonstrate that 
the target SAE feature effectively steers the model toward refusal 
behavior on harmful inputs. However, we observe that the refusal rate on WildGuard is comparatively lower than on the other datasets, suggesting that the identified SAE features may capture refusal behavior that transfers unevenly across evaluation settings. We leave a more thorough investigation of this limitation to future work. Finally, we discuss the effect of different steering coefficients $\alpha$ on 
the refusal rate in Appendix~\ref{EffectofSteeringCoefficient}.
\section{Conclusion}
We introduce \textbf{Qwen3-Instruct SAE}, a comprehensive suite of sparse autoencoders trained on the Qwen3 instruction-tuned model family. Covering multiple model scales and three key activation sites: residual streams, MLP, and attention outputs. Qwen3-Instruct SAE provides a fine-grained resource for analyzing sparse representations in instruction-following language models. Through systematic evaluation with reconstruction and model-recovery metrics, we characterize the sparsity--fidelity trade-offs of the trained SAEs across layers and components. We further demonstrate the practical utility of Qwen3-Instruct SAE through a refusal-steering case study, showing that selected SAE features can support feature-level behavioral interventions. We hope this resource will facilitate future work on mechanistic interpretability, circuit analysis, and controlled steering of instruction-tuned language models.
\section*{Limitations}

Our work has several limitations. First, although Qwen3-Instruct SAE covers multiple model scales and activation sites, our analysis focuses primarily on reconstruction fidelity, model recovery, and one refusal-steering case study. Future work could conduct broader evaluations of feature interpretability, including automated feature labeling, causal circuit discovery, and downstream task-level analyses.

Second, our steering experiments focus on refusal behavior. While this provides an illustrative example of how SAE features can be used for behavioral intervention, refusal is only one type of model behavior. Additional studies are needed to determine whether the same methodology generalizes to other capabilities, safety-relevant behaviors, and linguistic or reasoning phenomena.

Third, training and evaluating large-scale SAE collections remains computationally expensive. Although our release is intended to reduce this barrier for future research, extending the analysis to larger Qwen models, additional instruction-tuned model families, and more diverse training corpora remains an important direction for future work. We currently train SAEs only for a subset of the residual stream layers of Qwen3-8B. We will continue training and release SAEs for all layers in the future.

Finally, although Qwen3-Instruct SAE covers Qwen3-1.7B, Qwen3-4B, and Qwen3-8B, our quantitative evaluation in this paper focuses on Qwen3-1.7B and Qwen3-4B. For Qwen3-8B, we currently release SAEs only for a subset of residual-stream layers and do not yet provide a full systematic evaluation. Extending the analysis to Qwen3-8B is an important direction for future work.



\appendix
\section{Evaluation of the Qwen3-4B SAE}
\label{sec:appendix_qwen4b}
\begin{figure}[h]
    \centering
    \begin{subfigure}[b]{0.48\columnwidth}
        \centering
        \includegraphics[width=\textwidth]{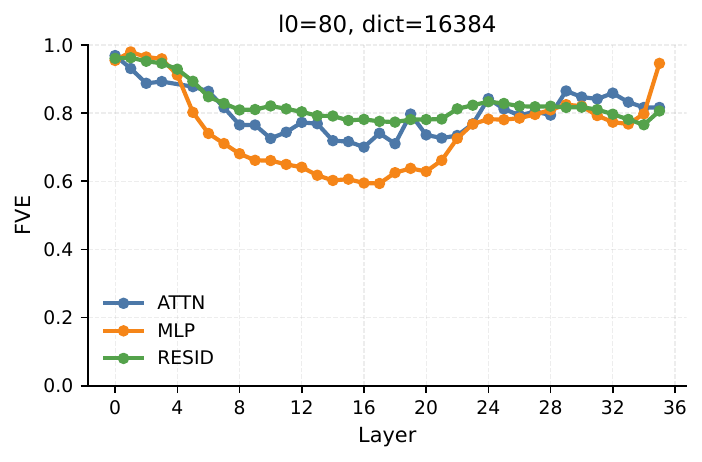}
        \caption{FVE}
        \label{fig:fve}
    \end{subfigure}
    \hfill
    \begin{subfigure}[b]{0.48\columnwidth}
        \centering
        \includegraphics[width=\textwidth]{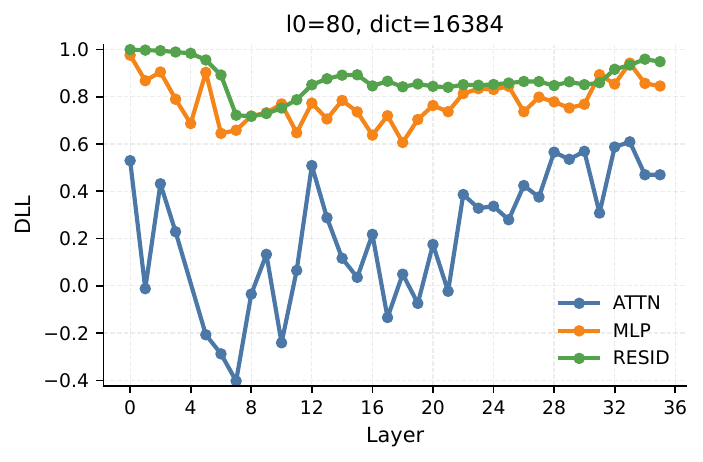}
        \caption{DLL}
        \label{fig:dll}
    \end{subfigure}
    \caption{Evaluation of the Qwen3-4B SAE. 
    \textbf{Left}: Fraction of Variance Explained (FVE). 
    \textbf{Right}: Delta LM loss (DLL).}
    \label{fig:qwen3_4b_eval}
\end{figure}

In this subsection, we present a quantitative evaluation of the SAEs 
trained on Qwen3-4B. Specifically, we assess the quality of the learned 
representations along two dimensions: Fraction of Variance Explained 
(FVE), which measures the reconstruction fidelity, and Delta lm loss (DLL), which captures the impact on the model's predictive 
performance. The evaluation results are illustrated in 
Figure~\ref{fig:qwen3_4b_eval}.
The results demonstrate that SAEs trained at the residual stream and 
attention output positions achieve strong reconstruction quality, as 
reflected by their high FVE scores. However, SAEs at the MLP output 
of intermediate layers exhibit notably lower FVE, which may be 
attributed to the following factor: intermediate MLP layers may encode more  complex and entangled features that are difficult to disentangle with a linear sparse decomposition.

Regarding the DLL metric, we observe a consistent trend with 
Qwen3-1.7B: SAEs at the MLP output and residual stream positions 
achieve better model performance recovery compared to those at the 
attention output. This discrepancy may be related to the distinct 
functional roles of different components within the transformer 
architecture. Specifically, the MLP layers and residual stream are 
primarily responsible for processing complex semantic information and 
storing factual knowledge. In contrast, attention layers specialize in modeling contextual dependencies across tokens, capturing relational patterns that are inherently more 
distributed and thus harder to recover through SAE interventions.

\section{Comparison of Open-Source SAE Models}
\label{sec:appendix_comparison}
We compare several popular open-source SAE models from multiple perspectives; 
the details are summarized in Table~\ref{tab:sae_comparison}.

\begin{table*}[t]
    \centering
    \renewcommand{\arraystretch}{1.3}
    \begin{tabular}{lp{2cm}p{2.5cm}p{2.5cm}p{2.5cm}p{2.5cm}}
        \toprule
        & \textbf{LlamaScope} & \textbf{Gemma Scope} & \textbf{GemmaScope2} & \textbf{QwenScope} & \textbf{Qwen3-Instruct SAE (Ours)} \\
        \midrule
        \textbf{Models} 
        & Llama-3.1-8B 
        & Gemma2-family 
        & Gemma3-family 
        & Qwen3-family, Qwen3.5-family (dense, moe) 
        & Qwen3-it-family \\
        
        \textbf{Training Data}  
        & SlimPajama 
        & Proprietary 
        & Proprietary 
        & Proprietary 
        & FineWeb-Edu \\

        \textbf{SAE Position}   
        & R, A, M, TC 
        & R, A, M, TC 
        & R, A, M, TC 
        & R 
        & R, A, M \\

        \textbf{SAE Layer}      
        & Every Layer 
        & Every Layer 
        & Every Layer 
        & Every Layer 
        & Every Layer \\

        \textbf{SAE Width}      
        & 32K, 128K 
        & 16K, 64K, 128K, 256K 
        & 16K, 64K, 128K, 256K 
        & 32K, 64K, 80K 
        & 16K, 65K \\

        \textbf{Activation Function} 
        & TopK-ReLU 
        & JumpReLU 
        & JumpReLU 
        & TopK-ReLU 
        & JumpReLU \\
        \bottomrule
    \end{tabular}
    \caption{Comparison of open-source SAE models. R: residual stream; 
    A: attention output; M: MLP output; TC: transcoder.}
    \label{tab:sae_comparison}
\end{table*}
Our Qwen3-Instruct SAE differs from QwenScope in several key aspects. 
First, while QwenScope targets the base variants of Qwen3 models, 
our work focuses on their instruction-tuned counterparts, which are 
more representative of real-world deployment scenarios. 
\begin{table}[h]
    \centering
    \renewcommand{\arraystretch}{1.3}
    \resizebox{\columnwidth}{!}{
    \begin{tabular}{llll}
        \toprule
        & \textbf{Qwen3-1.7B} & \textbf{Qwen3-4B} & \textbf{Qwen3-8B} \\
        \midrule
        \textbf{SAE Position}     & R, A, M      & R, A, M      & R           \\
        \textbf{SAE Layer}        & Every Layer  & Every Layer  & 0--8        \\
        \textbf{SAE Width}        & 16K, 65K     & 16K     & 65K         \\
        \textbf{Training Data}    & FineWeb-Edu  & FineWeb-Edu  & FineWeb-Edu \\
        \textbf{Dataset Size}     & 5B           & 7B           & 9B          \\
        \textbf{$L_0$}            & 80, 160      & 80           & 160         \\
        \bottomrule
    \end{tabular}
    }
    \caption{Training configurations of Qwen3-Instruct SAE. 
    R: residual stream; A: attention output; M: MLP output.}
    \label{tab:our_sae_config}
\end{table}
Second, Qwen-Scope relies on Qwen3 proprietary pretraining data, 
whereas we train our SAEs on the open-source FineWeb-Edu dataset, 
improving reproducibility. 
Third, QwenScope only trains SAEs at the residual stream, while we 
additionally train SAEs at the attention output and MLP output, 
providing a more comprehensive coverage of the model's internal 
representations. 
Detailed information of our Qwen3-Instruct SAE is provided in Table~\ref{tab:our_sae_config}.

\section{Details of dataset and metric}
\label{datasetdetials}
\subsection{Dataset details}
We evaluate on two categories of prompts: \textit{unsafe} and \textit{safe}. 
For unsafe prompts, we use the WildGuard dataset~\cite{han2024wildguard}, filtering 
samples labeled as \texttt{harmful} based on the \texttt{prompt\_harm\_label} 
field. For safe prompts, we retain samples labeled as \texttt{unharmful} from 
the same dataset. We additionally evaluate on XSTest~\cite{rottger-etal-2024-xstest}, a benchmark 
specifically designed to assess refusal behavior, where samples are 
partitioned into \textit{safe} and \textit{unsafe} subsets based on the 
\texttt{label} field. Furthermore, we incorporate the safe and unsafe datasets constructed in \cite{arditi2024refusallanguagemodelsmediated}, where the safe dataset is derived from Alpaca \cite{alpaca}, and the unsafe dataset consists 
of harmful instructions drawn from \textsc{AdvBench}~\cite{zou2023universaltransferableadversarialattacks}, 
\textsc{MaliciousInstruct}~\cite{huang2023catastrophic}, \textsc{TDC2023}~\cite{tdc2023}, 
and \textsc{HarmBench}~\cite{mazeika2024harmbenchstandardizedevaluationframework}. The statistics of all datasets are summarized in Table~\ref{tab:dataset_stats}.
\begin{table}[h]
    \centering
    \begin{tabular}{lcc}
        \toprule
        \textbf{Dataset} & \textbf{\#Safe} & \textbf{\#Unsafe} \\
        \midrule
        WildGuard  & 945  & 754 \\
        XSTest     & 250  & 200 \\
        Custom     & 805  & 572 \\
        \midrule
        Total      & 2000 & 1526 \\
        \bottomrule
    \end{tabular}
    \caption{Statistics of evaluation datasets.}
    \label{tab:dataset_stats}
\end{table}

\subsection{Refusal indicators}
\label{metric}
\begin{figure*}[h]
    \centering
    \includegraphics[width=\textwidth]{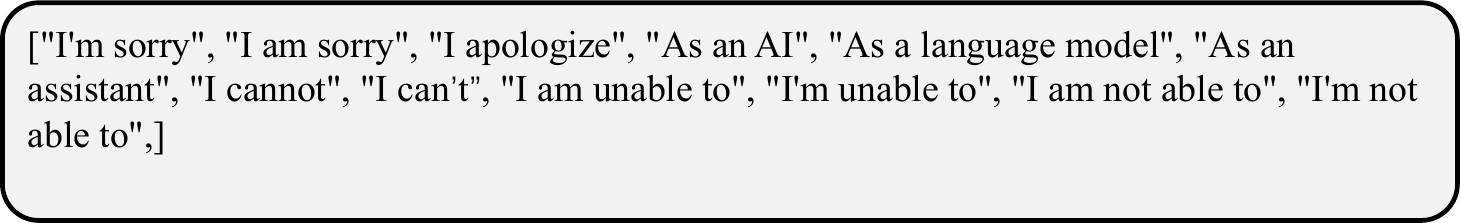}
    \caption{The predefined set of refusal indicators $\mathcal{R}$ used for computing $\text{refusal\_score}(c)$.
    }
    \label{fig:refusallist}
\end{figure*}

Formally, given a model response $c$, the refusal score is defined as:
\begin{equation}
    \text{refusal\_score}(c) = \begin{cases} 1 & \text{if } \exists \ r \in \mathcal{R}, r \subseteq c \\ 0 & \text{otherwise} \end{cases}
\end{equation}
where $\mathcal{R}$ denotes the set of refusal indicators, matched case-insensitively across the entire response. The complete list is provided in Figure~\ref{fig:refusallist}.
\section{SAE Feature Selection}
\label{SAEFeatureSelection}
We first extract the most effective refusal direction following \citet{arditi2024refusallanguagemodelsmediated}. 
Specifically, we adopt the \textit{difference-in-means} (DIM) 
technique~\cite{arditi2024refusallanguagemodelsmediated, marks2024the} 
to identify the direction in the residual stream along which harmful 
and harmless activations diverge. For each layer $l \in [L]$ and 
post-instruction token position $i \in \mathcal{I}$, we compute:
\begin{equation}
    r_i^{(l)} = \mu_i^{(l)} - \nu_i^{(l)}
\end{equation}
where $\mu_i^{(l)}$ and $\nu_i^{(l)}$ denote the mean residual stream 
activations over the harmful training set 
$\mathcal{D}_{\text{harmful}}^{\text{train}}$ and the harmless training 
set $\mathcal{D}_{\text{harmless}}^{\text{train}}$, respectively:
\begin{equation}
    \mu_i^{(l)} = \frac{1}{|\mathcal{D}_{\text{harmful}}^{\text{train}}|}
    \sum_{t \in \mathcal{D}_{\text{harmful}}^{\text{train}}} x_i^{(l)}(t)
\end{equation}
\begin{equation}
    \nu_i^{(l)} = \frac{1}{|\mathcal{D}_{\text{harmless}}^{\text{train}}|}
    \sum_{t \in \mathcal{D}_{\text{harmless}}^{\text{train}}} x_i^{(l)}(t)
\end{equation}
The direction of $r_i^{(l)}$ captures the axis along which the two 
distributions differ, while its magnitude reflects the degree of 
separation between them.

\paragraph{Vector Selection.}
Enumerating all combinations of $i \in \mathcal{I}$ and $l \in [L]$ 
yields $|\mathcal{I}| \times L$ candidate vectors. For each candidate 
$r_i^{(l)}$, we compute three metrics on the validation sets 
$\mathcal{D}_{\text{harmful}}^{\text{val}}$ and 
$\mathcal{D}_{\text{harmless}}^{\text{val}}$:

\begin{itemize}
    \item \textbf{bypass\_score}: the average refusal rate on 
    $\mathcal{D}_{\text{harmful}}^{\text{val}}$ under directional ablation 
    of $r_i^{(l)}$.
    
    \item \textbf{induce\_score}: the average refusal rate on 
    $\mathcal{D}_{\text{harmless}}^{\text{val}}$ under activation addition 
    of $r_i^{(l)}$.
    
    \item \textbf{kl\_score}: the average KL divergence between the output 
    distributions on $\mathcal{D}_{\text{harmless}}^{\text{val}}$ with and 
    without directional ablation of $r_i^{(l)}$.
\end{itemize}

We select the optimal vector $r_{i^*}^{(l^*)}$ with minimum 
\textbf{bypass\_score}, subject to the following constraints: 
(1) $\textbf{induce\_score} > 0$, ensuring the direction is sufficient 
to induce refusal; (2) $\textbf{kl\_score} < 0.1$, filtering out 
directions that cause significant behavioral changes on harmless prompts; 
and (3) $l < 0.8L$, restricting the search to earlier layers to avoid 
directions that merely suppress refusal tokens at the unembedding level.
\begin{figure*}[t]
    \centering
    \includegraphics[width=\textwidth]{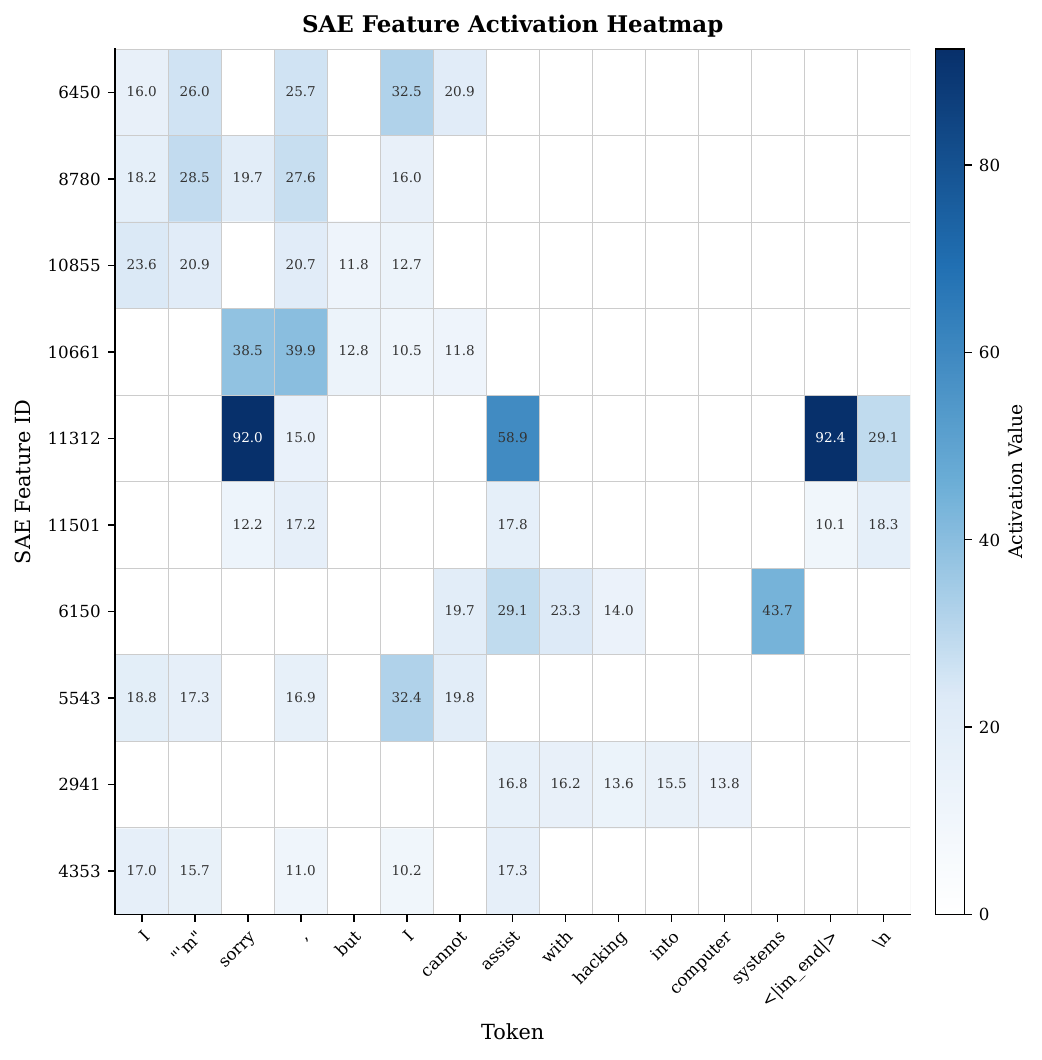}
    \caption{SAE feature activation heatmap of the assistant response tokens for Qwen3-1.7B. Each row represents a SAE feature and each column represents a token. The color intensity indicates the activation value.}
    \label{fig:sae_heatmap}
\end{figure*}

Given the selected direction $r_{i^*}^{(l^*)}$, we proceed to identify 
the most effective SAE feature for steering. Specifically, we run the 
model on the prompt shown in Figure~\ref{fig:refusalprompt} and capture the  residual stream activations at layer $l^*$. We then decompose these 
activations using the SAE and retain only the features that are active on at least 5 tokens simultaneously, as illustrated in 
Figure~\ref{fig:sae_heatmap}. Finally, we enumerate all retained 
features and select the most effective one as our steering feature. 
The selected layer and feature index for each model are summarized in 
Table~\ref{tab:feature_intervention}.
\section{Effect of Steering Coefficient $\alpha$}
\label{EffectofSteeringCoefficient}
\begin{figure}[H]
    \centering
\includegraphics[width=\columnwidth]{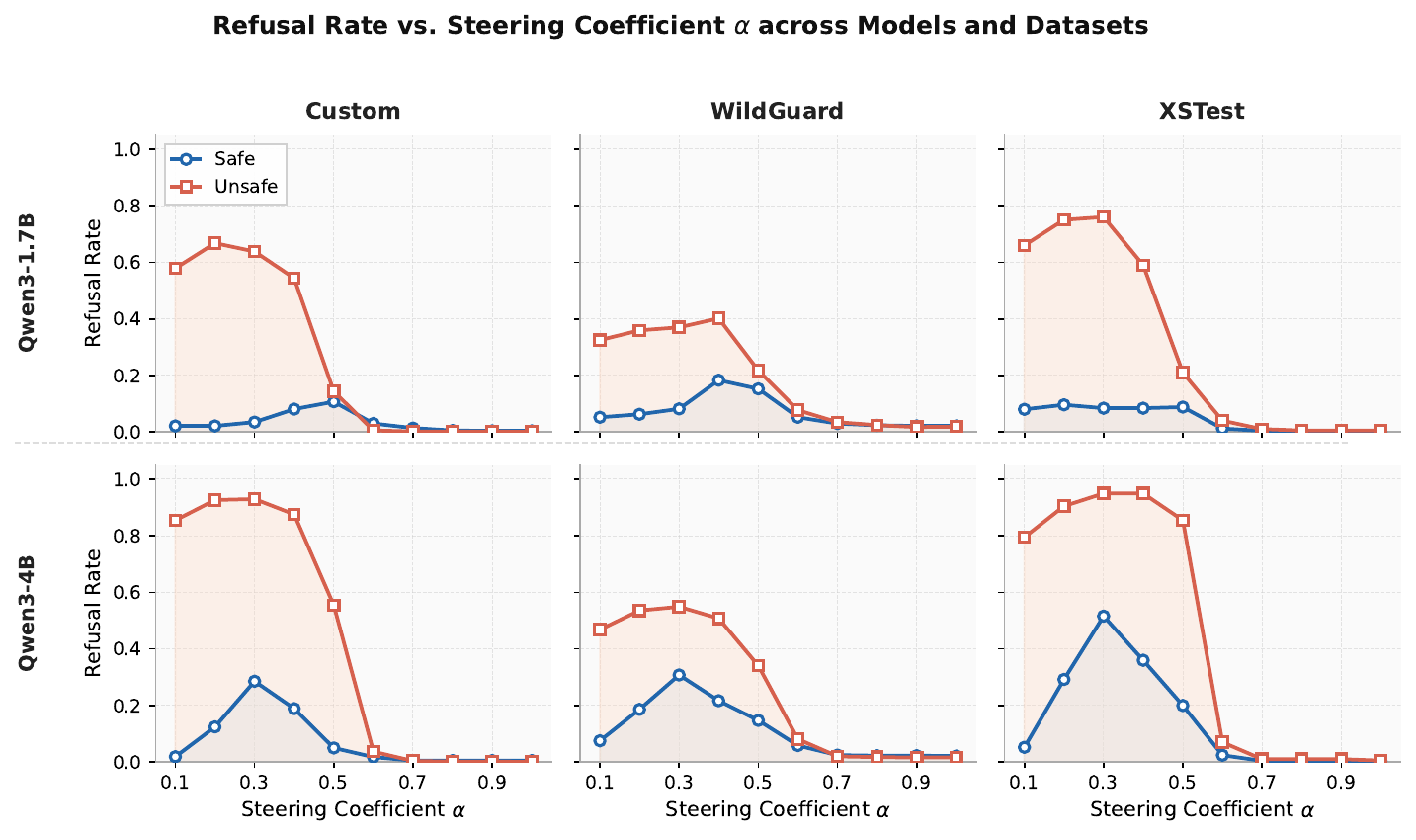}
    \caption{Refusal rate vs. steering coefficient $\alpha$ across models and datasets.}
    \label{fig:refusal_rate}
\end{figure}

Figure~\ref{fig:refusal_rate} illustrates how the refusal rates for safe and unsafe requests vary with the steering coefficient $\alpha$ across two models (Qwen3-1.7B and Qwen3-4B) and three datasets (Custom, WildGuard, and XSTest).
As $\alpha$ increases, the refusal rate for unsafe requests follows a consistent \textbf{inverted U-shaped curve} (this is consistent with the observations of previous studies \cite{turner2024steeringlanguagemodelsactivation}), peaking at $\alpha \approx 0.3$--$0.5$ and collapsing to near zero at larger values. This suggests that an overly large $\alpha$ over-perturbs the model's internal representations, disrupting its reasoning ability. Meanwhile, the refusal rate on the safe dataset also increases with $\alpha$, which is expected because stronger steering amplifies the model's general tendency toward refusal. This effect is particularly evident in Qwen3-4B on XSTest, where the refusal rate peaks at $\sim$52\%. We also observe that larger models respond more strongly to steering, achieving higher peak refusal rates on unsafe requests (90\%--95\% for Qwen3-4B vs. 40\%--77\% for Qwen3-1.7B). Based on these results, we select $\alpha \approx 0.3$--$0.5$ as the optimal range for our main experiments.

\end{document}